# DashCam Video: A complementary low-cost data stream for on-demand forest-infrastructure system monitoring

Durga Joshi [1*], Chandi Witharana[1], Robert Fahey[1], Thomas Worthley[1], Zhe Zhu[1], Diego Cerrai[2]

[1] Department of Natural Resources and the Environment, Eversource Energy Center, University of Connecticut, Storrs, CT 06269, USA

[2] Department of Civil and Environmental Engineering, Eversource Energy Center, University of Connecticut, Storrs, CT 06269, USA

*Author to whom correspondence should be addressed.

## Abstract

Urban green infrastructure (UGI) plays a critical role in enhancing ecological resilience and reducing infrastructure vulnerability in metropolitan settings. However, achieving scalable, high-resolution monitoring of UGI remains a persistent challenge due to visual occlusion, structural complexity, and the cost or inaccessibility of conventional 3D remote sensing technologies. Our study introduces a novel, low-cost, and reproducible framework for real-time, object-level structural assessment and geolocation of roadside vegetation and infrastructure using commonly available but underutilized dashboard camera (dashcam) video data. We developed an end-to-end pipeline that combines monocular depth estimation, depth error correction, and geometric triangulation to generate accurate spatial and structural data from continuous street-level video streams acquired from vehicle-mounted dashcams. Depth maps were first estimated using a state-of-the-art monocular depth model, then refined via a gradient-boosted regression framework to correct underestimations, particularly for distant objects. The depth correction model achieved strong predictive performance ($R^2 = 0.92$, MAE = 0.31 on transformed scale), significantly reducing bias beyond 15 m. Further, object locations were estimated using GPS-based triangulation and frame-by-frame image angles, while object heights were calculated using pin hole camera geometry. Our method was evaluated under varying conditions of camera placement and vehicle speed. The configuration involving interior-mounted cameras and low-speed travel yielded the highest accuracy, with mean geolocation error of 2.83 meters (IQR = 2.64 m), and mean absolute error (MAE) in height estimation of 2.09 m for trees and 0.88 m for poles. To the best of our knowledge, this is the first framework to combine monocular depth modeling, triangulated GPS-based geolocation, and real-time structural assessment for urban vegetation and infrastructure using consumer-grade video data. Our approach complements conventional overhead remote sensing methods, such as LiDAR and stereo imaging by offering a fast, real-time, and cost-effective solution for object-level monitoring of vegetation risks and infrastructure exposure, making it especially valuable for municipalities, utility companies, and urban planners aiming for scalable and frequent assessments in dynamic urban environments.

## 1.1 Introduction:

Urban landscapes are dynamic and complex socio-ecological systems characterized by the co-location of natural elements and anthropogenic infrastructure. Within these systems, urban green infrastructure (UGI), such as parks, urban forests, and street trees, serve as a critical component of urban resilience and sustainability. UGI enhances the quality of life by providing essential ecosystem services (e.g., air quality improvement, pollutant filtration, noise reduction, temperature regulation, and recreation) (Cadenasso & Pickett, 2000; Jin et al., 2014; Ow & Ghosh, 2017; Salmond et al., 2016; Smith, 2012; Tong et al., 2015; Weber et al., 2014). The environmental, social, and economic benefits derived from UGI are well established, yet their integration into densely built environments introduces a range of operational and planning challenges (Cook et al., 2024).

Among various UGI types, street trees serve a functionally vital role, though they remain relatively underrepresented in urban ecological research. The vegetated corridors, comprised of street trees, shrubs, sidewalks, and residential zones, offer substantial ecosystem value. However, they are often spatially constrained and directly interface with critical infrastructure, such as roads, buildings, utility poles, and powerlines (Coder, 1998; Randrup et al., 2001). Inadequate tree management, especially for trees in proximity to infrastructure, may introduce cascading risks, such as structural failure of trees during extreme weather events, disruption of utility services, impaired sightlines for traffic, and pedestrian safety hazards (Fröhlich et al., 2024; Wedagedara et al., 2023). To effectively address these challenges, it is essential to do quantitative assessment of tree structural parameters (e.g., crown geometry, stem diameter at breast height (DBH), health metrics) and their spatiotemporal proximity to roadside infrastructure. It is imperative for informed vegetation management, adaptive risk mitigation and resource allocation through vegetation management strategies.

Effective UGI management, mainly in transportation corridors, necessitates integrated spatial frameworks that incorporate both vegetation characteristics and urban infrastructure. Conventional UGI monitoring relies on field-based tree inventories that have intrinsic limitations in scalability, temporal resolution, and spatial granularity. Manual surveys are labor-intensive, cost-prohibitive for municipal-scale applications, and prone to data obsolescence in rapidly urbanizing regions. Consequently, urban planners and municipal agencies face significant challenges in maintaining up-to-date and spatially detailed datasets, limiting their ability to proactively manage vegetation–infrastructure conflicts. Recent advancements in remote sensing and geospatial technologies have significantly enhanced our ability to monitor and manage urban green spaces (Kowe et al., 2021; Nawar et al., 2022). Satellite and aerial platforms have been widely used in large-scale urban space assessments (Xu et al., 2020; Zhu, Z. et al., 2019). However, they fall short of providing detailed ground-level information in real time for precise and timely vegetation management. Even though very high-resolution images often offer individual tree-level information, their utility might be impacted by factors, such as shadows, viewing angle, occlusion, and seasonal variability. While advanced technologies, such as airborne LiDAR, hyperspectral imaging, and mobile laser scanning (MLS), offer enhanced spatial coverage, their operational costs, spectral resolution constraints, and logistical complexity hinder routine deployment for roadside vegetation monitoring (D'hont et al., 2024). This creates a critical data gap, impeding the development of predictive models for vegetation-infrastructure interaction dynamics and evidence-based urban forestry policies.

In remote sensing there are two fundamental challenges that arise as system complexity increases: the sensory gap and the semantic gap (Grubinger, 2007). The sensory gap arises during data acquisition (limitations such as spatial, temporal, and spectral) and is mainly associated with the sensor or platform used. The semantic gap emerges during data interpretation (human vs. machine) and may misinterpret the features and their relationship in the complex environment (Figure 1). Consequently, sensor and semantic gaps contribute to reducing the overall effectiveness and interpretability of remote sensing in understanding the existing scenarios. The level of abstraction we seek in an image by means of object boundary, object class, or both might not be able to achieve just based on manual, semi-automated, or fully automated approaches (e.g., rule-based, fuzzy, statistical, or machine learning) because if the sensor has not captured sufficient information there is always a compromise in the detections. While advancement in machine learning models (e.g. deep learning in image segmentation) has narrowed the semantic gap, the sensory gap, mainly associated with overhead remote sensing, often exacerbates the level of abstraction by affecting the quality and completeness of the existing information. Even at full operational capacity, sensors inherently lose information content of the real-world scenario and become increasingly pronounced as the complexity of the system increases. For instance, for relatively less complex systems roadside forest infrastructure, critical data gaps are due to the limitation of the sensors such as occlusion from canopy cover or insufficient spatial resolution. Overhead sensors can capture the extent of the canopy but are unable to capture the sub-canopy thus losing the vertical depth. Oblique sensors or on-ground sensors may fill the unique niche but may not capture the spatial depth. These compounded limitations further exacerbate the semantic information loss hence minimizing the achievable level of abstraction in data interpretation. Figure 1 shows the mechanisms underlying these sensory and semantic gaps.

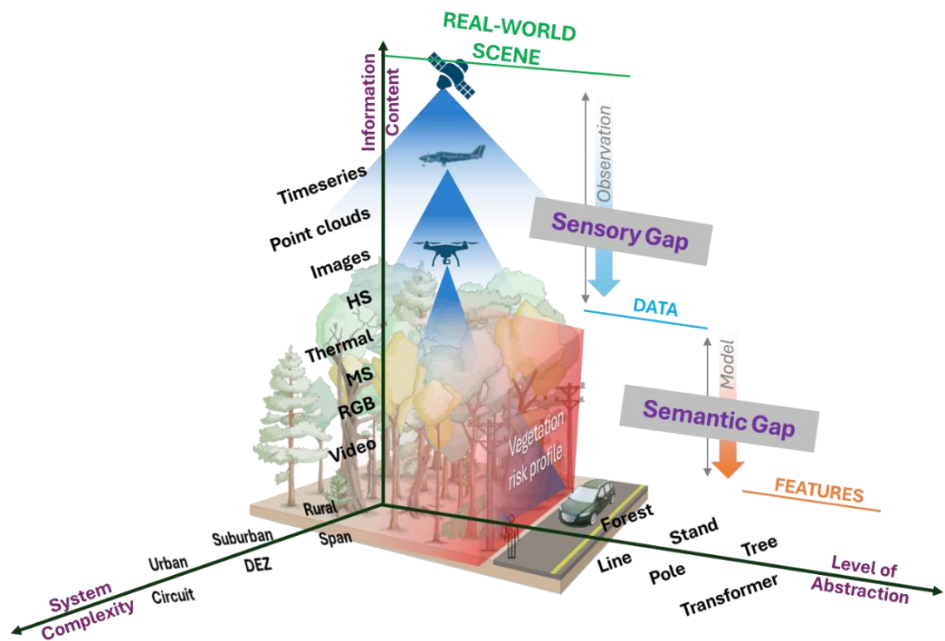

*Figure 1: Conceptual illustration of the sensory and semantic gaps in remote sensing that hinder the accurate translation of physical environments into geospatial information.*

Emerging advances in opportunistic remote sensing (ORS) and computer vision offer novel opportunities to address limitations associated with the sensor gaps by leveraging non-traditional data sources such as oblique perspectives from on-ground sensors. Vehicle-based remote sensing data such as Google Street View (GSV) and dashcam videos offer a complementary approach by capturing high-temporal-frequency, street-level imagery with inherent geospatial metadata (e.g., GPS location, acquisition date and time, and sensor geometry) of urban vegetation and infrastructure (Seiferling et al., 2017). Recent studies have demonstrated the utility of GSV images for urban vegetation analysis (Biljecki & Ito, 2021; Kim & Jang, 2023). The oblique-perspective information has revolutionized urban analytics by enabling large-scale, fine-grained observations of the built environment from a human-centric viewpoint (Carrasco-Hernandez et al., 2015; Gong et al., 2018; Li et al., 2018). Street-level imagery that mirrors pedestrian experiences has made it indispensable for deriving spatially explicit metrics, such as pedestrian visibility of greenery (Li et al., 2015), street canyon geometry (Hu et al., 2020), and microclimatic indicators like sky view factor (Liang et al., 2020). Even though GSV is spatially extensive, it exhibits temporal latency due to irregular update cycles (often spanning multiple years) which hinder dynamic monitoring of vegetation phenology, infrastructure interactions, and/or rapid urban land-use changes.

In recent years, vehicle-based data capture has undergone a dramatic transformation fueled by the widespread adoption of dashboard camera (dashcam) technology. These compact, low-cost cameras, mounted on dashboards or windshields, continuously capture high-resolution video footage of the surrounding environment. Dashcams can record a wide spectrum of contextual information on roadside environments, including diverse vegetation types, weather conditions, and traffic patterns. Dashcam videos have become popular in the transportation and engineering sector for characterizing and monitoring roads, bridges, traffic signs, and cracks (Dadashova et al., 2021; Hou et al., 2022). Dashcams are widely used in countries like Russia, Taiwan, and China for vehicle safety and surveillance (Rea et al., 2018), and in the U.S., companies such as Uber and Lyft employ them for liability and insurance purposes (Lyft, 2022; UBER, 2022). Modern vehicles, including Tesla with built-in multi-camera systems, now continuously capture high-resolution, georeferenced video, offering emerging opportunities for environmental and infrastructure monitoring, such as an assessment of roadside vegetation dynamics and utility infrastructure risk, beyond their original legal applications. One of the key advantages of dashcam systems lies in their ability to deliver a continuous stream of georeferenced, time-stamped visual data, which can be leveraged for real-time or retrospective automated analysis. With advances in computer vision and artificial intelligence, particularly deep learning (DL), it is now feasible to automatically process dashcam video feeds to extract relevant features. DL models can be trained to automatically extract critical information from dashcam videos, such as tree health and structure (height, crown size, leaning branches, DBH), to assess roadside vegetation health and potential hazards. (Joshi & Witharana, 2022) made the first effort to investigate the practicality of DL for classifying roadside vegetation using dashcam videos. Their work focused on differentiating between lower and higher vegetation types, employing image segmentation techniques to achieve this classification. These capabilities facilitate proactive vegetation management by enabling near real-time assessment of roadside vegetation dynamics, including encroachment, structural integrity, and spatial relationships with adjacent infrastructure.

While in-vehicle-based camera systems are increasingly leveraged for urban sensing applications, the use of dashcam video footage for detailed analysis of roadside vegetation and infrastructure remains largely unexplored. Existing studies have primarily focused on object detection and classification tasks within urban traffic environments (e.g.,(Boyd et al., 2022; Giovannini et al., 2021; Joshi & Witharana, 2022; Li, 2021; Taccari et al., 2018)), often overlooking the complex spatial relationships and structural variability inherent in roadside vegetation, especially in relation to adjacent infrastructure such as utility poles, powerlines, and roadways. This research gap is particularly pronounced in the context of vegetation monitoring, where occlusions, diverse plant forms, and varying light conditions present analytical challenges that are not well addressed by current street-level imaging approaches.

To the best of our knowledge, no prior studies have been conducted to explore the capability of consumer-grade dashcam videos to monitor roadside-forest infrastructure systems. This study seeks to address the limitations of existing remote sensing datasets by exploring the potential of dashcam video data as a resource for near real-time high-resolution, ground-level assessment of urban roadside vegetation and infrastructure. The approach enables detailed structural interpretation of vegetation and its spatial proximity to infrastructure elements, supporting applications in vegetation risk analysis, urban ecological monitoring, and spatial planning. Specific objectives of this study are to: (1) geolocate target objects along the urban corridor, and (2) estimate individual object-level structural metrics, such as height andwidth of target object. By focusing on precision and contextual interpretation, the methodology offers a transparent and reproducible means of extracting critical urban landscape information from an underutilized yet widely available data source. Furthermore, this study highlights both the opportunities and challenges of deploying dashcam-based systems in operational settings, including measurement errors, error propagation, uncertainty quantification, computational limitations, and constraints related to real-world implementation.

## 1.2 Methodology:

A generalized schematic of our modeling framework is shown in Figure 2. First, we decompose dashcam video footage into individual frames. Each frame is then processed through a monocular depth estimation model to generate depth maps, which provide pixel-wise distance estimates from the camera to objects within the scene. Simultaneously, metadata that comprises timestamps, and geographical coordinates are extracted from the video to temporally and spatially align each frame. These geotagged frames are used to associate depth measurements with known camera positions. Target objects, such as utility poles, trees, and roadside signs, are identified in the frames, and their dimensions (e.g., height and width) are measured in image space. For each object, depth values are sampled from multiple pixels and averaged to reduce noise and improve reliability. A triangulation approach, based on repeated observations of the same object from different camera positions (determined via GPS and timestamp metadata), is employed to estimate the object's geographic coordinates. To improve the accuracy of structural measurements, particularly for objects situated on sloped terrain, a high-resolution digital elevation model (DEM) was integrated into the analysis. The DEM allowed terrain elevation adjustments to be applied to the estimated object positions, resulting in more accurate height and crown width estimations. These corrected measurements were then cross-referenced with the field measurement to validate structural parameters. This integrated approach enables robust, GPS-supported object geolocation and structural characterization using only passive monocular video data, even in complex roadside environments with varying elevations.

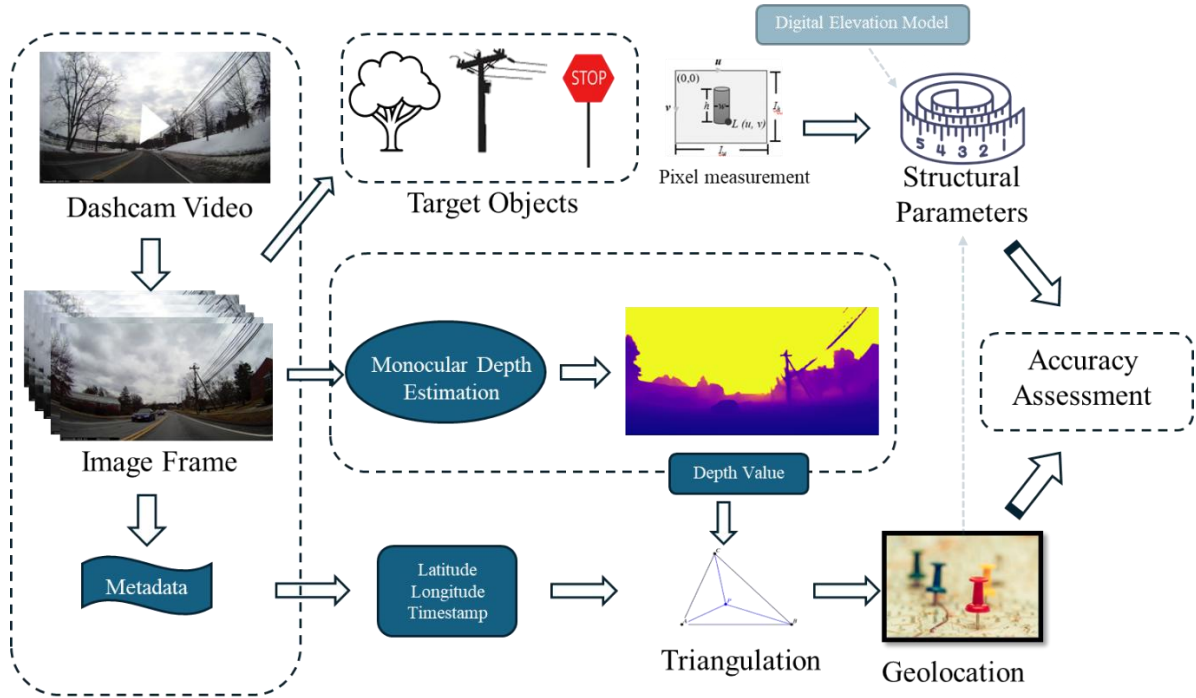

*Figure 2: Generalized Workflow for Monocular Depth-Based Object Geolocation and Structural characterization from Dashcam video frames*

### 1.2.1 Study Area & Data Collection:

We collected video data along a 3 km road circuit of Storrs Mansfield, Connecticut, USA (Figure 3). The circuit is characterized by diverse roadside features, including vegetation, utility poles, traffic signs, and various concrete structures with differing heights and spatial configurations. The terrain exhibited both upslope and downslope variations relative to the roadway, contributing to a complex spatial environment. A low-cost Thinkware U1000 dashcam mounted on a moving vehicle was used for data acquisition. This camera captures 4K-resolution video at 30 frames per second and is equipped with GPS functionality, enabling spatial tagging of video frames. The GPS data is collected every 1sec period (Table 1).

*Table 1: Table showing the specification of the dashboard camera (Dashcam) used in our study.*

| Dashcam | Properties |
|---|---|
| Name | Thinkware U1000 |
| Resolution @ frame per second (fps) | 4K UHD (3840 x 2160) @ 30 fps or 2K QHD (2560 x 1440) @ 60 fps |
| GPS | Enabled |
| GPS rate | 1sec |
| Field of view (FOV) (diagonal) | 152º |
| Focal Length | 3.6 mm |
| Lens | 8.5MP Sony Starvis IMX335 |
| Price | USD 349.99 |

For detailed evaluation, we focused on a representative 1 km segment of the road that includes multiple utility poles, trees, cross-arms, and transformers. Ground truth locations of these target objects were collected using GPS, and their structural parameters such as height, diameter at breast height (DBH), and crown width were measured manually.

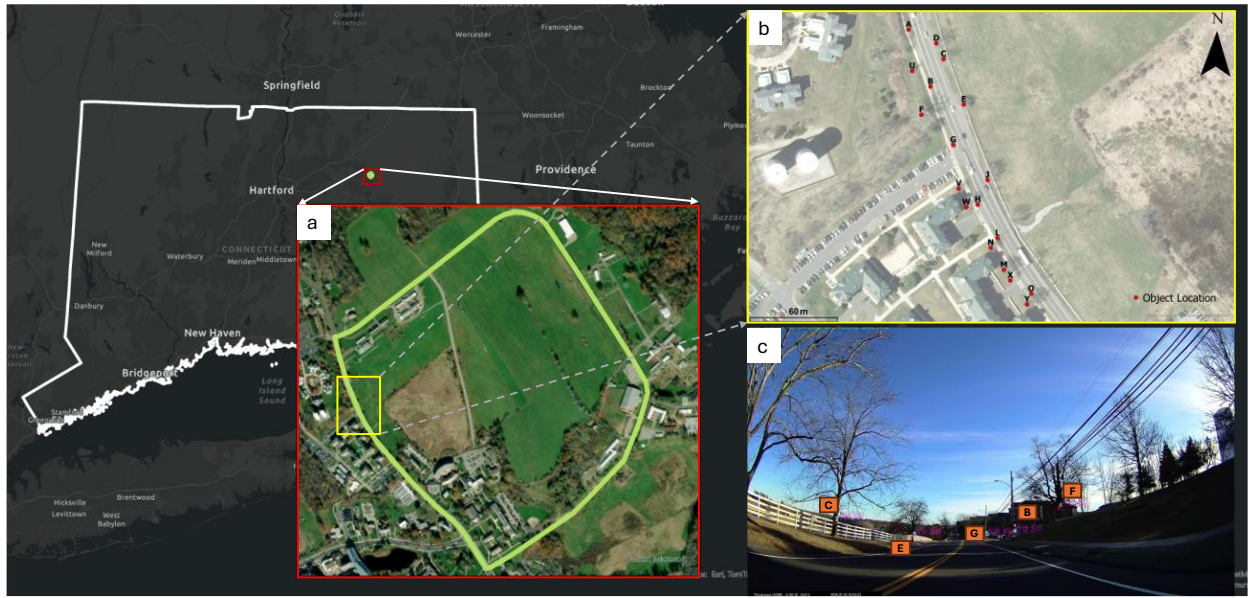

*Figure 3: (a) Map showing the 3 km loop area collected with dashcam video. (b) Aerial image showing the georeferenced object locations (red points) used for validation, with corrected positions indicated. (c) Street-level dashcam camera image with annotated objects used for depth estimation. For each object, depth values were extracted from three distinct pixels (denoted by small red dots) in the depth image and averaged.*

To assess the influence of vehicle speed and camera placement on measurement accuracy, the video data were collected under two operational speeds (below and above or equal to 40 km/h) and two camera placements (external and internal mounts). These configurations allowed us to evaluate the effect of motion blur and windscreen refraction on object geolocation accuracy, and structural information extraction. Because the dashcam embedded GPS coordinates at a 1 Hz sampling rate, we used every 30th frame with GPS metadata in the analysis to extract temporally aligned video frames for downstream analysis. The Thinkware U1000 dashcam exhibited a GPS positional uncertainty of approximately ±4 meters. The GPS error was largely systematic across frames, meaning that correcting the initial GPS position effectively aligned subsequent frames due to the consistent offset. To enhance spatial accuracy, we applied manual corrections using identifiable ground reference features. Our manual adjustments were guided by visual alignment with known landmarks and were performed within the bounds of the documented GPS uncertainty. These corrections were incorporated into the overall error budget and accounted for in the object triangulation and geolocation workflow.

### 1.2.2 Depth Estimation

Upon the completion of our dataset collection, we created depth images for each of the frames to extract depth information (Figure 4). We employed the Depth-anything V2 model (Yang et al., 2024)(Yang et al., 2024) to extract depth information from our dashcam video frames. Depth-anything V2 is the state-of-the-art monocular depth estimation model which has the hybrid architecture of Convolutional Neural Networks (CNNs) and transformer-based components that employ knowledge distillation. The model excels at estimating depth from single images due to its training on a large and diverse dataset that includes both real and synthetic imagery, which enables robust generalization to novel image types. However, as the model processes frame individually without incorporating temporal information, its predictions exhibit flickering i.e., temporal inconsistencies between adjacent frames due to its invariant training objectives. While developing or fine-tuning a depth model tailored to our dataset was beyond the scope of our study, we used the pre-trained model and subsequently scaled and converted the outputs into metric units.

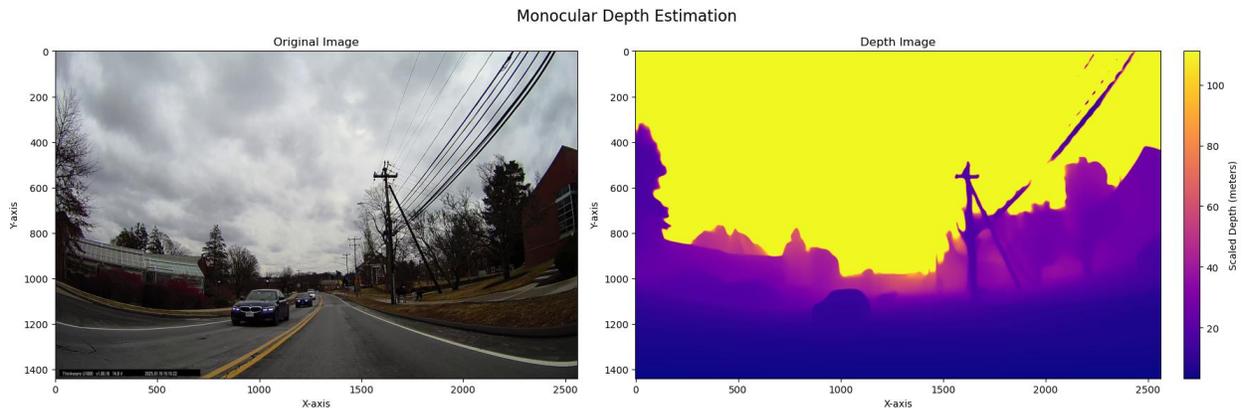

*Figure 4: Original RGB image showing roadside infrastructure including vehicles, utility poles, and vegetation. Right: Corresponding depth map derived from monocular depth estimation, with warmer colors (yellow) indicating greater distances and cooler colors (purple/blue) indicating closer objects. The depth values are scaled in meters.*

It is worth noting that, because the model was not fine-tuned on our specific training data, the estimated depths for distant objects often deviated from the actual ground truth distance. Initially, we manually determined the object distance from the camera location to the object using the ArcGIS Pro 'Measure' tool. To estimate depth, we sampled video frames at 1 Hz (i.e., every 30th frame) and converted each into a depth map using the Depth Anything V2 model. For each object visible in a given frame, depth values were extracted from three representative pixel locations and averaged to obtain a depth estimate from that particular camera location for the object. While some objects may have appeared in multiple frames, we did not perform object tracking or aggregate depth values across frames. Each measurement was treated independently, and all analyses were conducted at the frame level. By treating each frame independently, without aggregating depth values across time or tracking objects across frames, we minimized the risk of spatial or temporal autocorrelation affecting our depth analysis. The process was repeated for 27 objects across more than 70 camera locations (Figure 3(b, c)). Depth values below approximately 15 m were comparable, but objects beyond 15 m from the camera position were often underestimated, as shown in Figure 5. However, due to the high speed of the vehicle, capturing enough frames of the same object presented a significant challenge, specifically the need for at least three distinct frames within a 15 m range to enable effective triangulation.

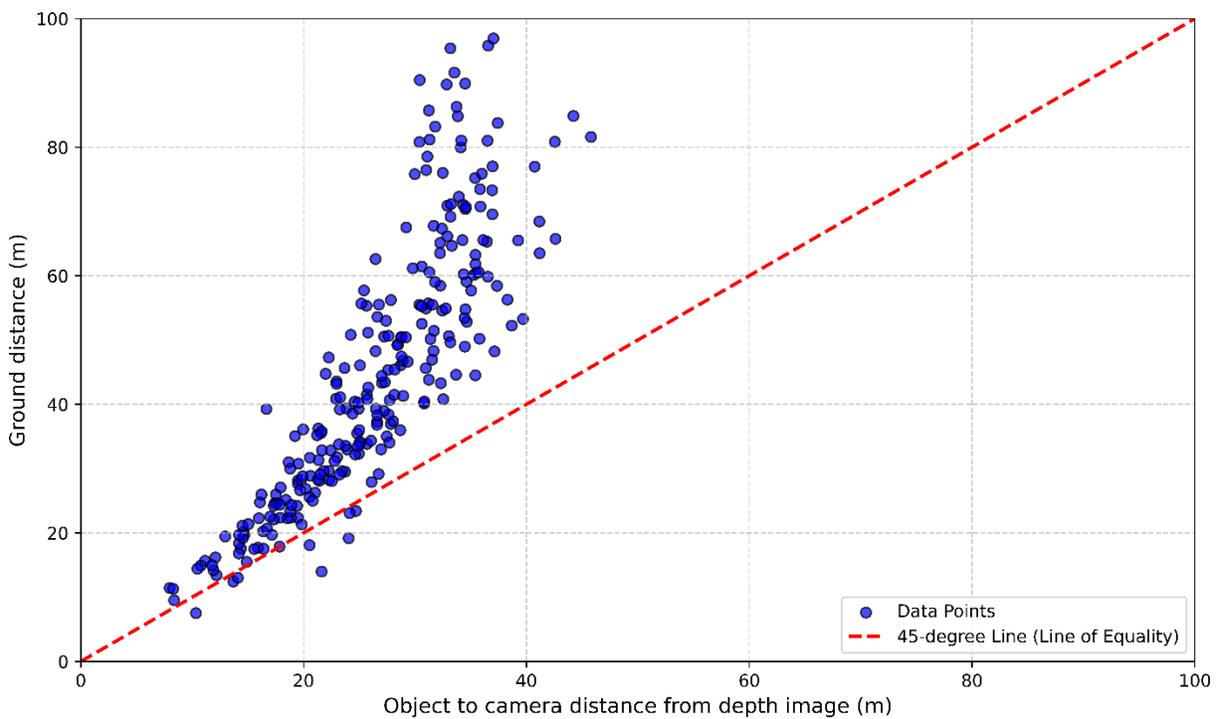

*Figure 5: Scatter plot showing the relationship between ground-measured object distances and corresponding distances estimated from monocular depth images. The red dashed line represents the 45-degree line of equality, indicating perfect agreement.*

To address the limitations of the predicted depth image, we used the XGBoost machine learning model (Brownlee, 2016) to predict the ground truth distance values for the distant object in the image frames. The model utilized various input variables to enhance the predictability of actual including the depth value at the pixel location, normalized pixel location for spatial information within the image, latitude and longitude of the image frame for geographical context, and the azimuth of the vehicle at the time the image was captured. The dataset was split into training (70%) and testing (30%) subsets based on individual video frames, ensuring that depth images used for training were not reused in testing. While the same physical object may appear in both sets, each appearance corresponds to a different frame captured from a distinct camera location, resulting in unique depth representations. Hyperparameters were optimized using the grid search method (Marinov & Karapetyan, 2019) and model selection were performed solely within the training set to mitigate overfitting. The model performance was evaluated using mean absolute error (MAE)(Chai & Draxler, 2014). After selecting the best hyperparameters, the final model was retrained on the full training set and evaluated on the held-out test set to assess generalization performance. The output of the model is the predicted distance from the camera to the object.

### 1.2.3 Geolocation

Geolocation of objects was achieved through a geometric median-based triangulation framework (Cohen et al., 2016), integrating multi-camera observations to estimate target coordinates (Figure 6). For each object $j$, candidate positions were generated by intersecting directional lines derived from depth $D_{(i,j)}$ and pixel-remapped angle $\beta_{(i,j)}$, where $i \in \{1,2,..., Nj\}$ denotes the sequential camera locations (or frames) observing object $j$, and $N_j$ is the total number of camera positions for object $j$. The angle $\beta_{(i,j)}$ was calculated by reprojecting pixel coordinates to real-world angles using the camera's intrinsic parameters (e.g., focal length, horizontal field of view (HFOV)). Spatial uncertainties in camera geolocation, depth estimation ($D_{(i,j)}$), and angular alignment ($\beta_{(i,j)}$) produced dispersed candidate points for each target. To resolve this dispersion, the geometric median of all candidate positions was computed, minimizing the sum of Euclidean distances to all points and yielding a robust estimate of the object's true location. The methodology involved preprocessing camera data, converting geographic coordinates to a local Cartesian system, generating directional lines representing object positions from each camera's perspective, and computing the geometric median of these lines to derive the final object location.

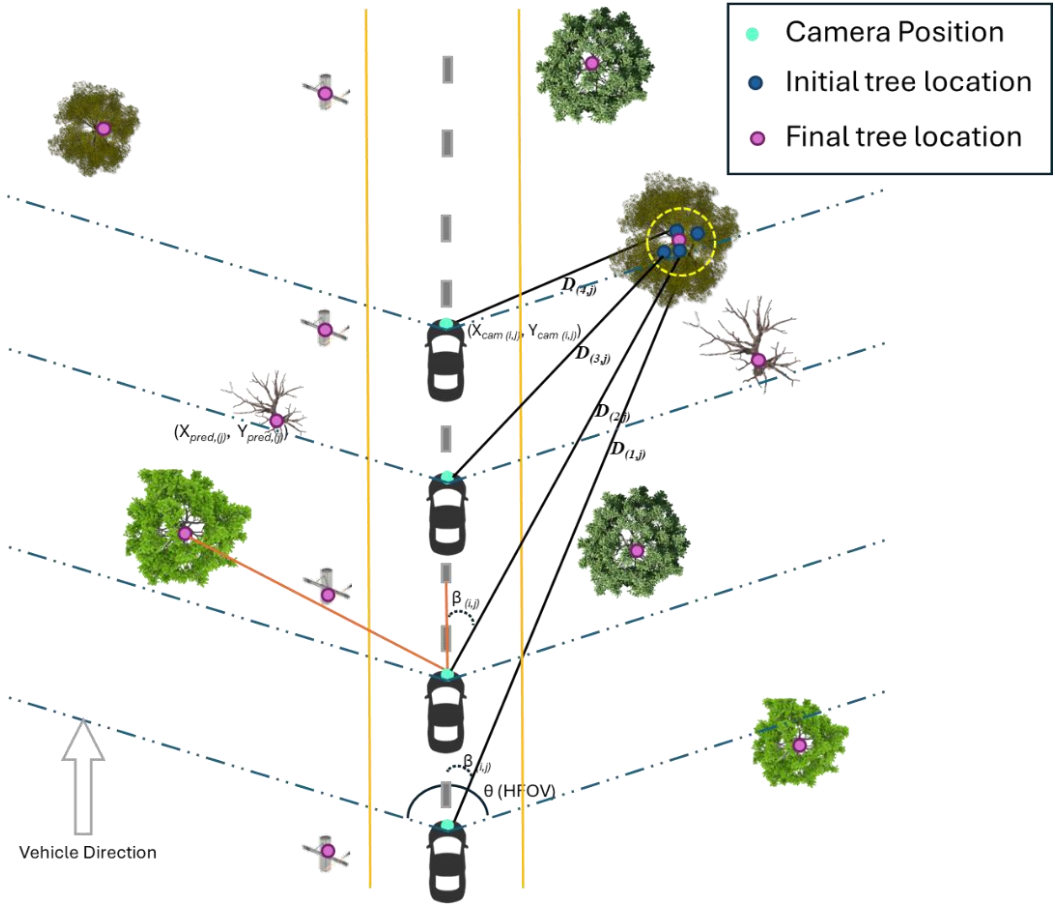

*Figure 6: A Schematic triangulation-based geolocation in roadside environments. The diagram illustrates a vehicle-mounted camera system capturing depth data along a roadway. Objects are located at varying distances from the road, and distances (d) are estimated from the camera to each object. Geolocation points (purple) are calculated using triangulation, based on the intersection of lines formed by the view angle (β) and distance (d) from multiple camera positions (denoted by cyan colors) observing the same object. The directional field of view of the camera is indicated by blue dashed lines, and vehicle movement is shown in the upward direction*

Data preprocessing began with the acquisition of camera parameters, including GPS coordinates (latitude and longitude) of camera positions ($X_{cam(i,j)}, Y_{cam(i,j)}$), pixel locations of objects within the images, depth estimates (distance from the camera to object), azimuth angles (camera orientation relative to true north), and pixel heights of objects. Pixel angles relative to the camera's centerline were calculated using the horizontal field of view (HFOV) and image dimensions. For a pixel location ($u, v$) in an image of width Iw, the angle $\theta$ was computed as:

$$\theta = -\left(\frac{u - I_w/2}{I_w}\right) \cdot HFOV \qquad (1)$$

where the negative sign accounts for the directionality of pixel displacement. Camera positions and object coordinates were converted from GPS to a local Cartesian coordinate system to simplify geometric calculations. We used reference points (the first camera's GPS position $(X_{cam(1,j)}, Y_{cam(1,j)})$, since $i = 1$ for first position) and predicted geographic coordinates onto a flat plane using Earth's radius (R = 6,378,137m) and trigonometric transformations. For a camera position $(X_{cam(1,j)}, Y_{cam(1,j)})$, the Cartesian coordinates $(X, Y)$ of an object $j$ with coordinate $(Xj, Yj)$ were derived as:

$$X = (X_j - X_{cam(1,j)}) \cdot \frac{\pi R}{180} \cdot \cos(Y_{cam(1,j)}) \quad (2)$$

$$Y = (Y_j - Y_{cam(1,j)}) \cdot \frac{\pi R}{180} \quad (3)$$

Directional lines representing object positions were generated for each camera view. Each line originated at the camera's Cartesian coordinates and extended along the calculated pixel angle $\beta_{(i,j)}$ for the corresponding depth $(d_i)$. The endpoint of the line was determined using trigonometric projections:

$$\Delta X = d \cdot \cos(\beta_{(i,j)}), \Delta Y = d \cdot \sin(\beta_{(i,j)}) \quad (4)$$

The geometric median of all line endpoints was computed to estimate the object's location. The geometric median minimizes the sum of Euclidean distances to all endpoints, providing robustness to outliers and measurement errors. This was formulated as an optimization problem:

$$geometric\ median\ (P^*) = \underset{p}{argmin} \sum_{i=1}^{n} ||P - P_i||_2 \quad (5)$$

Where $P_i$ are the endpoints of the lines, and $P$ is the estimated object location. The optimization was performed using the L-BFGS-B algorithm (Zhu, C. et al., 1997), initialized with the mean value of all endpoints to ensure convergence. Finally, the estimated Cartesian coordinates were converted back to GPS coordinates using inverse transformations. This approach ensured robustness to noisy data and provided a scalable solution for multi-camera geolocation tasks.

### 1.2.4 Object height and width estimation

To estimate the real-world height of objects from images, we use principles derived from the pinhole camera model, which describes how a 3D object is projected onto a 2D image plane (Figure 7). This relationship is grounded in the concept of similar triangles, which relate the real-world height of an object ($H$) to its projection on the camera sensor. The fundamental principle in an ideal pinhole camera is that the ratio of an object's real-world height $H$ to its distance from the camera $D$ is equal to the ratio of the object's height on the image sensor $H_{sensor}$ to the focal length $f$, expressed mathematically as:

$$\frac{H_{sensor}}{f} = \frac{H}{D} \quad (6)$$

To use the height measured in pixels h on the image, it must first be converted into its physical equivalent on the sensor, $H_{sensor}$. This is achieved using the relation:

$$H_{sensor} = h \cdot \left(\frac{S}{I_h}\right) \quad (7)$$

Where $\left(\frac{S}{I_h}\right)$ represents the physical size of each pixel on the sensor. Substituting this expression for $H_{sensor}$ into the earlier equation and rearranging yields the equation:

$$H = \frac{h \cdot D \cdot S}{f \cdot I_h} \quad (8)$$

Where, $H$ represents the real-world height of the object, $h$ is the object's height in the image measured in pixels, $D$ is the distance from the camera to the object (derived from depth estimation), $S$ is the physical height of the camera sensor (in units consistent with the focal length), $f$ is the focal length of the camera, and $I_h$ is the total height of the image in pixels. This method accounts for critical camera parameters (e.g., focal length, sensor size, and image resolution) and the object's distance from the camera to convert a pixel-based measurement into a real-world dimension. It relies on precise calibration of the camera and accurate depth estimation to ensure that the computed height is reliable and accurate. We used the same concept for the width estimation of the object using the formula where height variables are substituted with the corresponding width variable:

$$W = \frac{w \cdot D \cdot S}{f \cdot I_w} \quad (9)$$

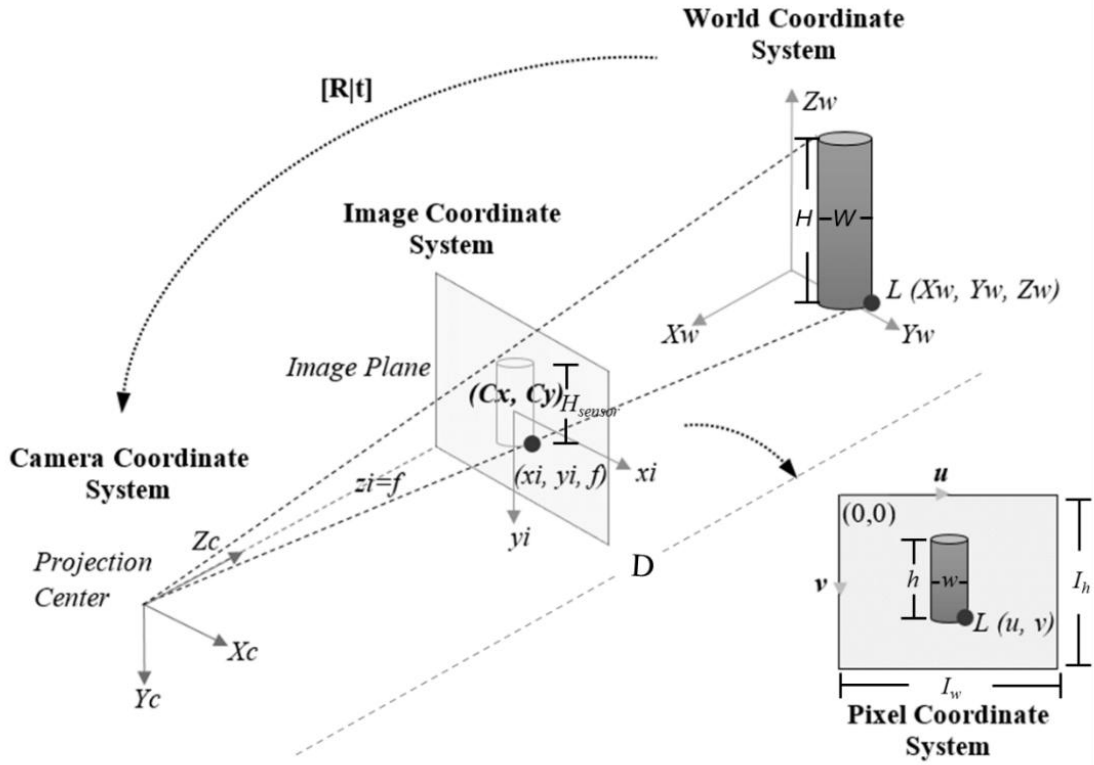

*Figure 7: Relationship Between World, Camera, Image, and Pixel Coordinate Systems in a Monocular Imaging Geometry. The diagram illustrates the transformation process from a 3D object point in the world coordinate system (Xw,Yw,Zw) to its corresponding 2D projection (u,v) in the pixel coordinate system. The object's real-world dimensions (height H and width W) are projected through the camera's intrinsic and extrinsic parameters. The projection involves a transformation from the world coordinate system to the camera coordinate system via rotation and translation [R|t], followed by perspective projection onto the image plane defined in the image coordinate system (xi,yi). This is then mapped onto the pixel coordinate system, with the camera's principal point (Cx, Cy), focal length f, and sensor height $H_{sensor}$ to determine the projected size and location of the object in image space.*

In scenarios where objects are located on elevated surfaces (e.g., roadside banks), the base of the object does not lie on the ground plane, violating the assumptions of the standard pinhole camera model. To address this, Digital Elevation Model (DEM) data was integrated into the pipeline. The base elevation of the object ($h_{elevation}$) at the predicted location was obtained from the DEM. The object's height ($H$) was calculated using the pinhole camera model, and the total height ($H_{total}$) was computed by adding the DEM-derived elevation. With the terrain slope ($\theta$), the horizontal distance ($D_{horizontal}$) was computed using the DEM-derived slope, and the corrected distance was used in the height estimation formula.

$$\beta = \arctan\left(\frac{h_{object} - h_{camera}}{D_{horizontal}}\right) \quad (10)$$

$$D_{horizontal} = \sqrt{D^2 - (h_{object} - h_{camera})^2} \tag{11}$$

$$\Delta H = H \cdot \tan(\beta) \tag{12}$$

$$H_{total} = H + \Delta H \tag{13}$$

Where $D$ is the distance from the camera to the object (derived from depth estimation), $f$ is the focal length of the camera, $\beta$ is the terrain slope angle derived from the DEM and $h$ is the height of the object in pixels.

### 1.2.5 Analysis

#### 1.2.5.1 Experimental Setup

Geolocation accuracy and prediction performance were evaluated through a controlled field experiment designed to quantify the influence of vehicle speed and camera mounting position (i.e. inside (on the windshield) or outside (on the hood)). Data were collected under four distinct operational conditions: the camera was mounted inside the vehicle at a height of 1.2 m and outside the vehicle at 0.9 m, while the vehicle was driven at both low speeds (below 40 km/h) and high speeds (above km/h). A total of 63 distinct objects (Table 2) were systematically selected within a consistent survey area to ensure repeated observation across all test conditions. These objects predominantly included utility poles, trees, traffic signals, and various forms of human-made infrastructure.

*Table 2: Distribution of objects included in the evaluation dataset, categorized by object type.*

| Object | Count |
|---|---|
| Tree | 38 |
| Utility Pole | 17 |
| Other | 8 |
| **Total** | **63** |

For each object, manual measurement on the image frame was performed to collect information such as depth values from the depth image and pixel level measurement from the video image frame. In the case of trees, both crown width and trunk diameter were extracted by measuring the pixel width at the crown and trunk levels, respectively. For linear structures, such as utility poles, lamps, traffic signs, and concrete structures, the pixel width at a representative location was recorded to estimate object width.

Geolocation error was defined as the geodetic distance between the predicted object position derived from our analysis and the ground-truth (GT) location. This error was computed using the haversine distance formula defined by the equation below,

$$d = 2R \cdot \arcsin\left(\sqrt{\sin^2\left(\frac{\emptyset_2 - \emptyset_1}{2}\right) + \cos(\emptyset_1)\cos(\emptyset_2)\sin^2\left(\frac{\lambda_2 - \lambda_1}{2}\right)}\right) \tag{14}$$

where d is the distance between two coordinates, R is the radius of the earth, Ø is latitude in radians and λ is longitude in radians.

*1.2.5.2 Statistical Analysis*

To evaluate the impact of camera position and vehicle speed on geolocation error, we conducted a paired statistical analysis for the experiment. Each object was observed under four scenarios: a camera mounted inside or outside the vehicle and driven at low (< 40km/h) or high (≥40 km/h) speeds. Given the non-normal distribution of errors and the paired structure, non-parametric tests were applied. The Wilcoxon signed-rank test assessed the effect of camera position (inside vs. outside) and vehicle speed (slow vs. speed) on the geolocation error. Additionally, a Friedman test was used to detect overall differences across the four scenarios. Also, we conducted pairwise Wilcoxon tests to further isolate speed effects within each camera position and tested camera position effects at each speed level. The Friedman test provided an omnibus test of condition-level effects, while pairwise Wilcoxon tests were additionally used to directly compare speed effects within each camera position and to compare inside versus outside camera positions at each speed level. To assess how geolocation error varies with object distance from the camera location, we grouped objects based on their distance from the camera in the final visible frame. Objects were binned into three distance categories: 0–10 m, 10–20 m, and 20–30 m. Geolocation errors were then computed and compared across these bins under all experimental conditions. We also assessed the relationship between geolocation error and object distance by calculating the correlation between geolocation error and the object's distance from the camera in both the first and last frames in which the object was visible. This analysis aimed to determine whether object distance influenced localization accuracy.

Prior to selecting the optimal acquisition condition, we conducted a comparative analysis of structural parameter estimation errors i.e., height, diameter at breast height (DBH), and crown width, across all four experimental scenarios. Based on our overall evaluation, we selected the scenario that yielded the lowest overall error as the best-performing configuration. Within this optimal scenario, we performed a detailed object-wise analysis of structural parameter estimation, stratified by object class: Tree, Pole, and Other (e.g., traffic signs, streetlights, and roadside concrete structures). For each object, we computed the absolute error in height, DBH, and crown width, allowing us to assess class-specific performance in structural estimation accuracy.

## 1.3 Results:

### 1.3.1 Depth Estimation with XGBoost Model:

The XGBoost model's performance was rigorously evaluated using both transformed and original scales to ensure robust predictions of ground measurements. Incorporating a square root transformation and variance-based weighting significantly enhanced the model's ability to handle non-linear relationships and varying data scales, resulting in stable performance across diverse terrain and sensor conditions. The model was tuned for hyperparameters and achieved a mean cross-validation mean absolute error (MAE) of 0.3965 across 10 folds, highlighting its consistency and reliability. On the transformed scale, the model demonstrated strong predictive accuracy, with a mean squared error (MSE) of 0.1797, an MAE of 0.3118, and a coefficient of determination ($R^2$) of 0.9200. When reverting to the original scale, the model showed its robustness (MSE of 36.5503, an MAE of 4.1590, and an $R^2$ of 0.9051), explaining over 90% of the total variation in the ground measurements.

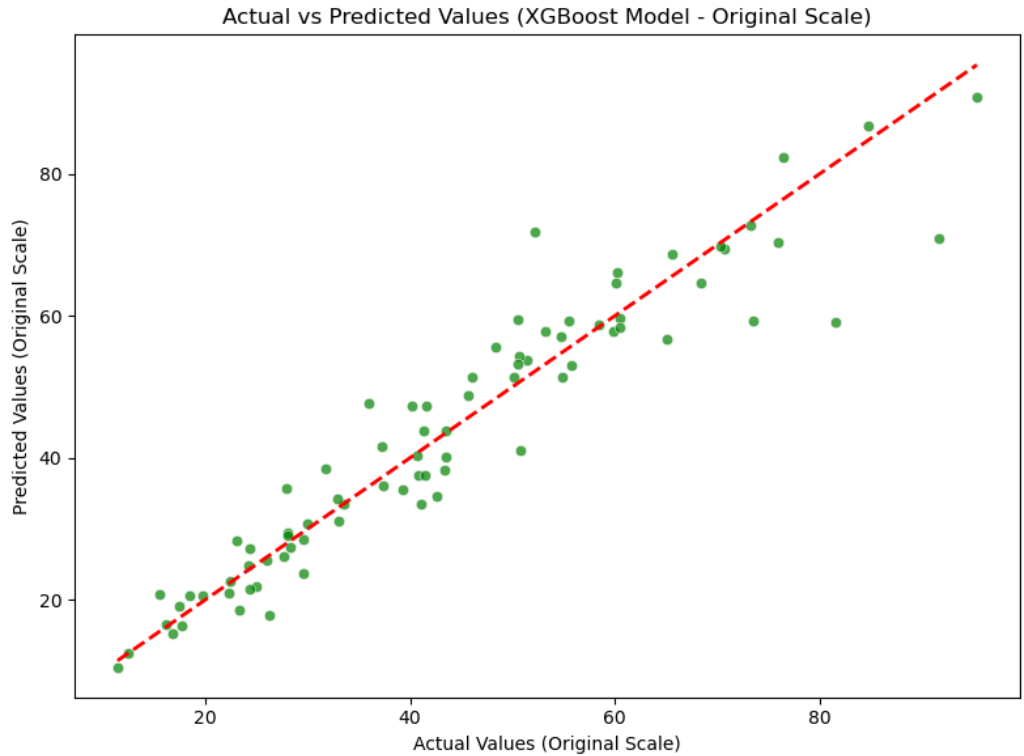

*Figure 8: Scatter plot of actual versus predicted values from the XGBoost regression model, evaluated on the original (untransformed) scale. Each green dot represents an individual observation, where the x-axis corresponds to the ground truth (actual) value, and the y-axis denotes the model's predicted value. The red dashed line represents the 1:1 line (perfect prediction).*

The use of the square root transformation and variance-based weighting significantly improved the model's ability to handle the non-linear relationships and varying scales within the data, leading to reduced prediction errors. The low MAE and high $R^2$ values across both scales and the data points closely aligned along the identity line (red dashed line) for the original scale (Figure 8) indicate a strong agreement between the actual and predicted values. With the model, we were able to effectively translate the image values to ground measurements with minimal bias. By coupling domain-aware transformations with uncertainty-guided weighting, the model achieved meter-level accuracy without requiring post-hoc corrections, making it suitable for real-time geolocation tasks in environmental monitoring and autonomous navigation applications.

### 1.3.2 Geolocation

The geolocation errors across each experimental condition are summarized in Table 3. The camera position (inside vs outside) and the speed of the vehicle influence the error produced. High-speed conditions consistently produced larger errors than slow-speed scenarios, with the mean error increasing by 77% for inside camera mounting and 36% for outside high-speed compared to their slow-speed counterparts. Inside measurements exhibited greater variability, particularly at high speeds, as evidenced by the 1.7 times wider interquartile range (IQR) for inside versus outside cameras positioned at high-speed conditions. However, in all four scenarios, the mean geolocation error is within the 5m radius range.

*Table 3: Summary statistics of measurement error (absolute value) across four experimental conditions based on platform location (Inside vs. Outside) and operational speed (Slow vs. High). The error represents the absolute difference between predicted and ground truth measurements for geolocation.*

| Condition | n | Mean | Median | STD | Min | 25 % | 75 % | Max | IQR |
|---|---|---|---|---|---|---|---|---|---|
| Inside and Slow speed (In_Slow) | 63 | 2.83 | 1.92 | 2.71 | 0.31 | 1.06 | 3.70 | 16.19 | 2.64 |
| Inside and High speed (In_Speed) | 62 | 5.01 | 3.60 | 4.26 | 0.15 | 1.68 | 7.94 | 16.57 | 6.25 |
| Outside and Slow speed (Out_Slow) | 63 | 3.04 | 2.24 | 2.53 | 0.48 | 1.43 | 3.82 | 10.93 | 2.39 |
| Outside and High speed (Out_Speed) | 62 | 4.13 | 3.44 | 3.14 | 0.47 | 1.86 | 5.50 | 16.12 | 3.63 |

#### 1.3.2.1 Prediction accuracy:

*Table 4: Summary of non-parametric statistical tests evaluating the effects of camera position and speed on measurement error across experimental scenarios. Wilcoxon Signed-Rank tests were used for paired comparisons, and a Friedman test was applied for overall within-subject differences across all four scenarios (In_Slow, In_Speed, Out_Slow, Out_Speed).*

| Test | Comparison | Statistic | p-value |
|---|---|---|---|
| Wilcoxon Signed-Rank | Camera Position (Inside vs Outside) | W = 919.0 | 0.6869 |
| Wilcoxon Signed-Rank | Speed Effect (Slow vs Fast) | W = 353.0 | <0.0001 |
| Friedman Test | Overall (All 4 Conditions) | $Chi^2$ = 19.29 | 0.0002 |
| Paired Wilcoxon | In_Slow vs In_Speed | W = 375.0 | <0.0001 |
| Paired Wilcoxon | Out_Slow vs Out_Speed | W = 488.0 | 0.0006 |
| Paired Wilcoxon | In_Slow vs Out_Slow | W = 739.0 | 0.0959 |
| Paired Wilcoxon | In_Speed vs Out_Speed | W = 778.0 | 0.1640 |

With the statistical analysis on different pairs, results showed that vehicle speed had a significant impact on geolocation error ($p = 1.20 \times 10^{-5}$), with higher speeds increasing the error regardless of camera position (Figure 9). In contrast, the camera mounting position did not significantly influence the error ($p = 0.686$).

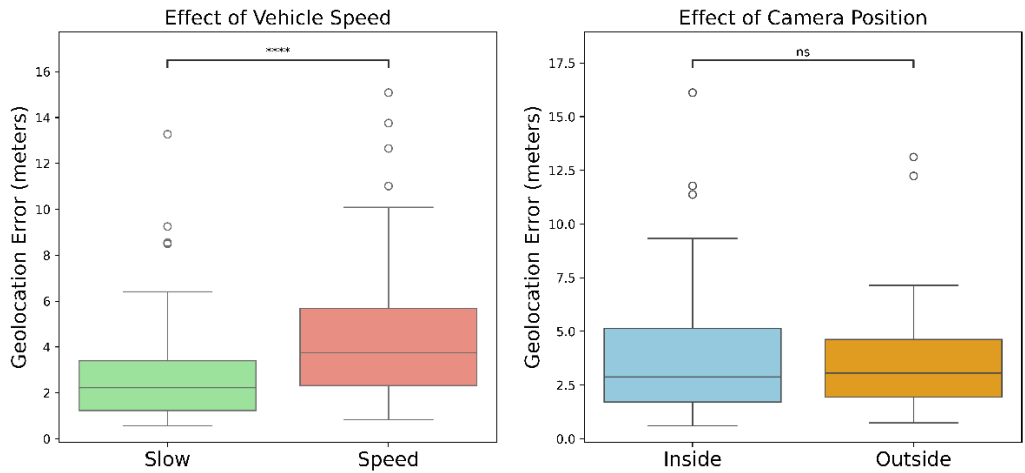

*Figure 9: Boxplots showing the influence of (a) vehicle speed and (b) camera mounting position on geolocation error (in meters). Each box represents the interquartile range with the median line, whiskers denote 1.5× IQR, and individual points indicate outliers. Statistical annotations above the plots reflect results from Wilcoxon Signed-Rank tests.*

Additionally, a Friedman test examined overall differences across all four experimental conditions and showed significant differences in error distributions ($\chi^2 = 19.29$, $p = 2.37 \times 10^{-4}$). Furthermore, pairwise Wilcoxon signed-rank tests showed that speed significantly increased geolocation error both inside (W = 375.0, $p = 2.5 \times 10^{-5}$) and outside (W = 488.0, $p = 6.15 \times 10^{-4}$) the vehicle (Table 4). However, no significant differences were observed between camera positions when compared at the same speed level (Figure 9), with In_Slow vs. Out_Slow (W = 739.0, p = 0.096) and In_Speed vs. Out_Speed (W = 778.0, p = 0.164) both being non-significant. These results indicate that vehicle speed is the dominant factor influencing geolocation accuracy, while the camera's mounting position (inside vs. outside) had minimal impact under the tested conditions.

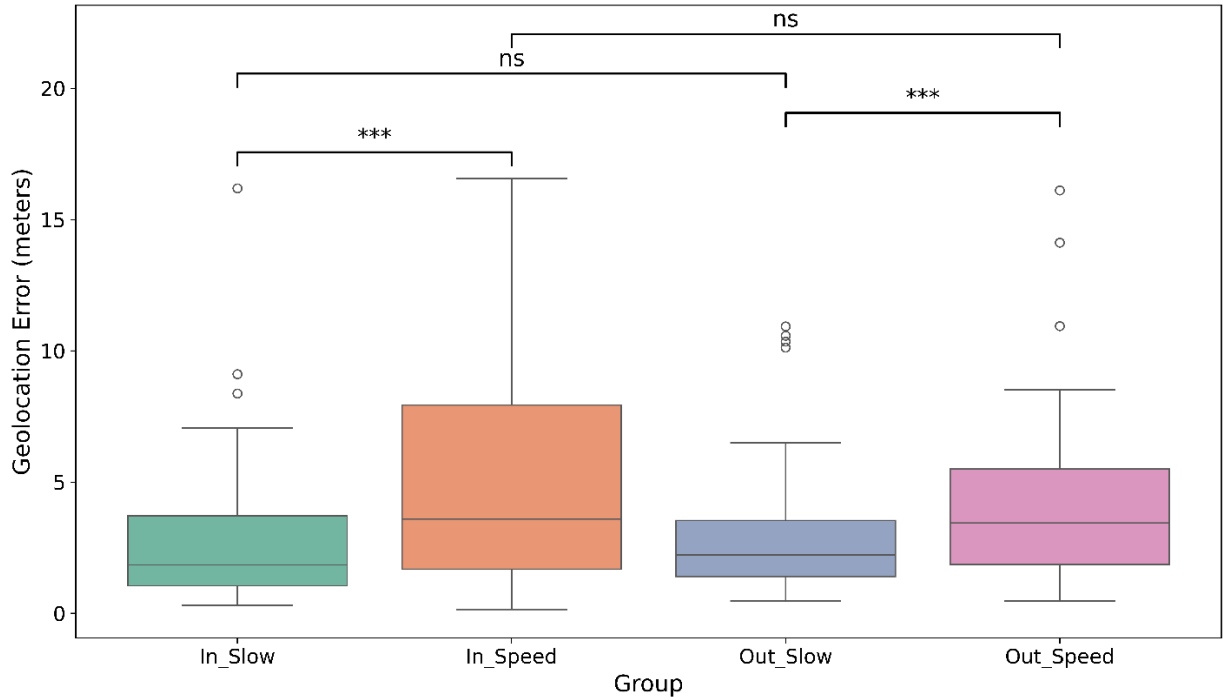

*Figure 10: Boxplots of geolocation error (in meters) across four experimental groups: In_Slow, In_Speed, Out_Slow, and Out_Speed. Boxes represent the interquartile range (IQR) with the median line, whiskers extend to 1.5× IQR, and outliers are shown as individual points. Pairwise statistical comparisons were performed using Wilcoxon Signed-Rank tests, with significance levels indicated: \*\*\*p < 0.001; ns = not significant.*

#### 1.3.2.2 Correlation Between Camera-Object Distances and Error

The relationship between the predicted object locations and the camera's spatial position during object visibility was evaluated using Spearman's rank correlation coefficient ($\rho$). Specifically, the distances from both the first and last camera frames in which each object appeared were compared with the predicted object location to assess the consistency of geolocation accuracy under varying experimental conditions.

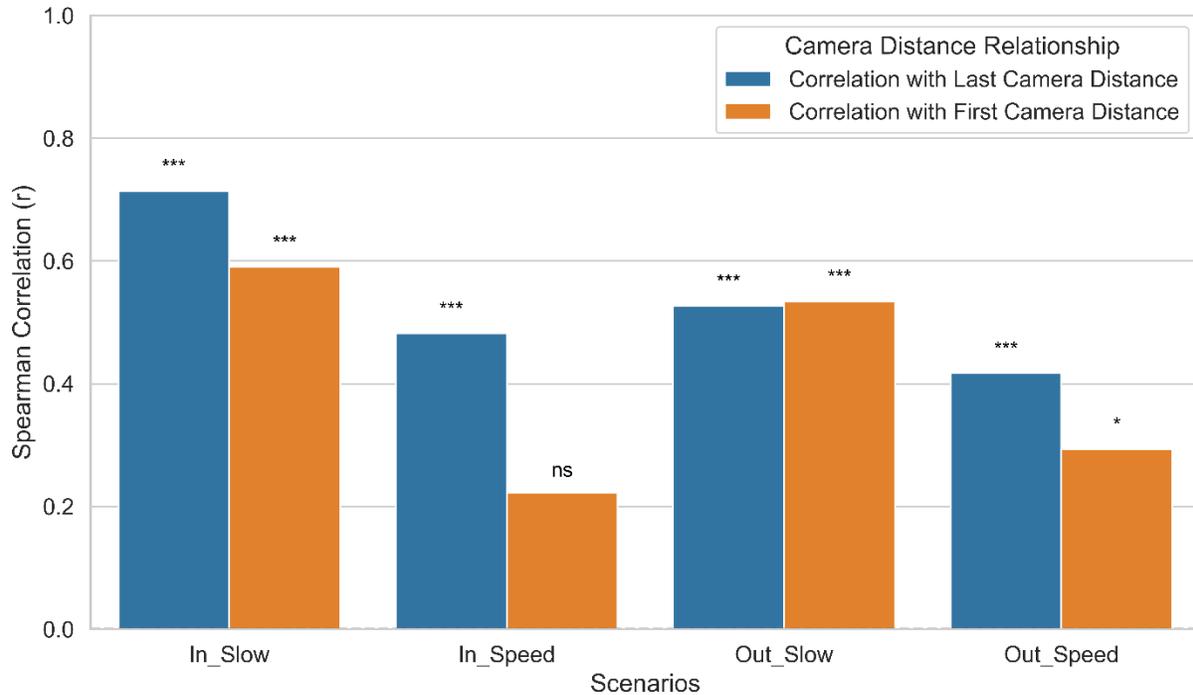

*Figure 11: Spearman's rank correlation (ρ) between geolocation error and measurements from the distance from the first and last camera location to the Ground truth location of the observed object across four experimental conditions: In_Slow, In_Speed, Out_Slow, and Out_Speed. Blue bars represent correlations with the distance from the last camera location, while orange bars represent correlations with the first. Asterisks indicate statistically significant correlations (\*\*\*p < 0.001, \*p < 0.05, ns=non-significant.)*

As illustrated in Figure 11, the last camera frame distance consistently exhibited a stronger correlation with the predicted object locations than the first camera frame distance across all four conditions. This pattern was particularly evident in the In_Slow condition, where Spearman's ρ value reached 0.72 for the last camera frame, indicating a high level of predictive reliability when the vehicle operated at slower speeds with the camera mounted inside. In contrast, the In_Speed and Out_Speed conditions demonstrated notably reduced correlation values, particularly when using the first camera frame distance.

Statistical significance, using spearman rank correlation test, further confirmed that these correlations were significant ($p < 0.05$), as denoted by asterisks in Figure 11. The results underscore that both vehicle speed and camera placement critically influence geolocation prediction accuracy, with slower speeds and interior camera mounting providing superior alignment between observed object distances and predicted locations.

### 1.3.2.3 *Impact of Object Distance on Geolocation Error*

The analysis revealed a clear and consistent trend in which geolocation error increased as the distance between the camera and the object grew (Figure 12).. For objects located within 0 -10m of the last camera frame, the geolocation error remained relatively low, with median values consistently below 2m across all conditions. This finding suggests that close-range objects are localized with high accuracy, regardless of vehicle speed or camera placement.

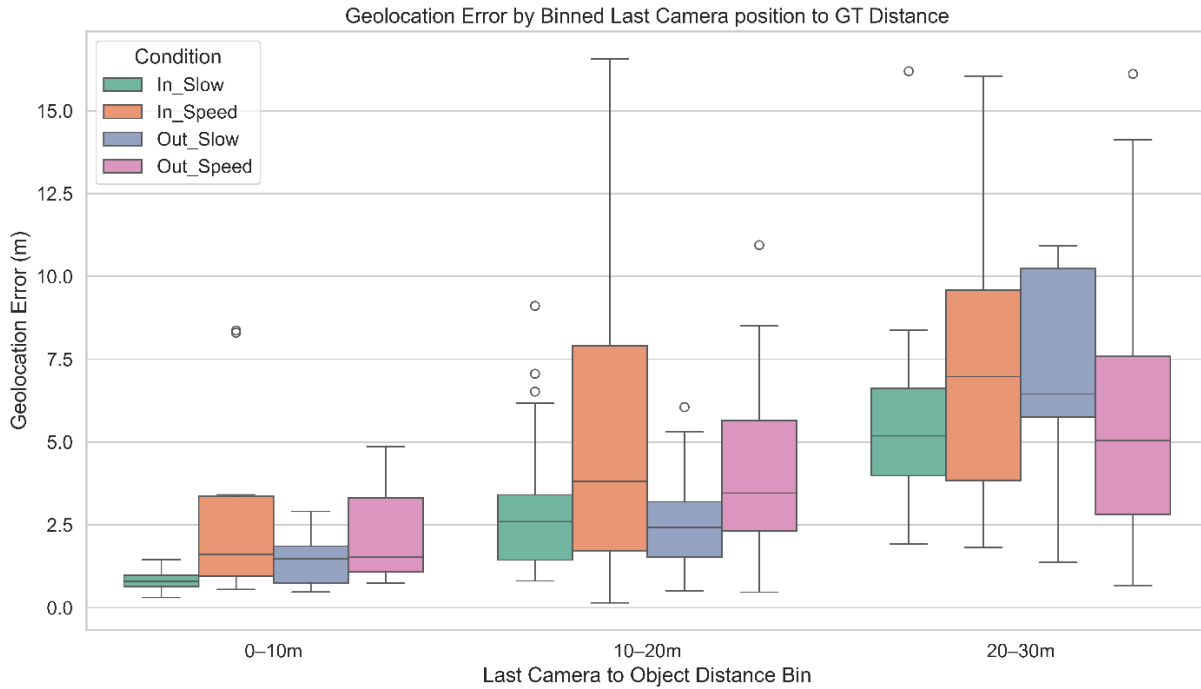

*Figure 12: Boxplots showing geolocation error (in meters) as a function of the last camera-to-object distance, binned into three intervals: 0–10 m, 10–20 m, and 20–30 m. Results are grouped by condition: In_Slow, In_Speed, Out_Slow, and Out_Speed. The plot illustrates the relationship between geolocation error and increasing distance from the last camera location of the observed object under varying speed and camera position scenarios.*

However, as the distance increased to the 10 m–20m bin, a noticeable rise in geolocation error was observed, particularly in the In_Speed condition, where error variability increased substantially, and several outliers exhibited errors reaching up to 17m. The 20–30m distance bin exhibited the highest geolocation errors across all conditions, with the most pronounced errors observed in the Out_Slow and In_Speed scenarios. These findings indicate that geolocation accuracy diminishes significantly for objects that are farther away from the camera, reflecting the increasing difficulty of precise localization as visual information degrades with distance.

Overall, the results demonstrate that geolocation performance is optimized when objects are within proximity to the camera and when the vehicle operates at slower speeds with the camera positioned inside the vehicle. Conversely, high-speed conditions and increased object distance exacerbate localization errors.

### 1.3.3 Structural Parameters:

The evaluation of structural errors (height, DBH and crown classes) per object class across four data collection scenarios (In_Slow, In_Speed, Out_Slow, Out_Speed) revealed significant differences in attribute estimation performance. Wilcoxon signed-rank tests with Bonferroni correction demonstrated distinct error patterns across scenarios, with particularly notable differences in height and crown width estimation.

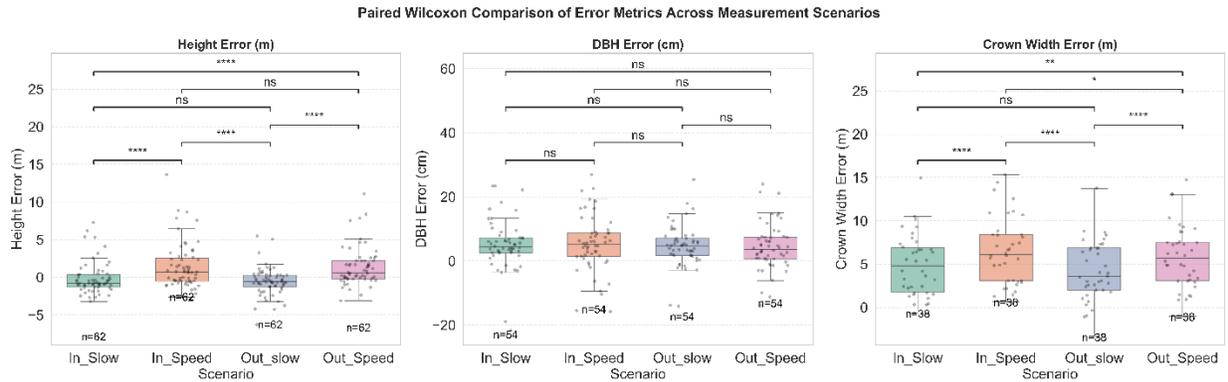

*Figure 13: Paired Wilcoxon comparison of error metrics across four measurement scenarios: In_Slow, In_Speed, Out_Slow, and Out_Speed. Boxplots show the distribution of errors in (a) height (m), (b) DBH (cm), and (c) crown width (m). Statistical significance of pairwise comparisons is annotated above each plot using Wilcoxon Signed-Rank tests (ns = not significant; p < 0.05 (\*), 0.01 (\*\*), 0.001 (\*\*\*), 0.0001 (\*\*\*\*)). Sample sizes (n) are shown below each group.*

For height estimation, In_Slow scenario demonstrated the lowest mean height error (-0.24 m) and median error of -0.76 m (IQR=1.69), significantly lower than In_Speed (median= 0.70 m, IQR= 3.09; $p=4.28 \times 10^{-7}$) and Out_Speed (median= -0.58 m, IQR=2.41; $p=7.93 \times 10^{-7}$). The Wilcoxon tests revealed that In_Slow and Out_slow showed comparable height accuracy (p=1.0).

*Table 5: Summary statistics and pairwise Wilcoxon Signed-Rank test results for measurement error across four experimental scenarios (In_Slow, In_Speed, Out_Slow, Out_Speed) for three tree attributes: height (m), DBH (cm), and crown width (m). The table reports the sample size (N), mean signed error (Mean), Mean absolute error (MAE) median, standard deviation (STD), and interquartile range (IQR) for each scenario. P-values reflect comparisons against the In_Slow baseline using paired Wilcoxon tests, with significance levels denoted as follows: ns = not significant; p < 0.05 (\*), 0.01 (\*\*), 0.001 (\*\*\*), 0.0001 (\*\*\*\*)*

| Error | Scenario | N | Mean | MAE | Median | STD | IQR | p-value (vs. In_Slow) | Significance |
|---|---|---|---|---|---|---|---|---|---|
| Height (m) | In_Slow | 62 | -0.24 | 1.66 | -0.76 | 2.36 | 1.69 | - | - |
|  | In_Speed | 62 | 1.45 | 2.19 | 0.70 | 3.08 | 3.09 | $4.275 \times 10^{-07}$ | **** |
|  | Out_Slow | 62 | -0.58 | 1.38 | -0.61 | 1.90 | 1.54 | 1.0000 | ns |
|  | Out_Speed | 62 | 1.27 | 1.97 | 0.58 | 2.77 | 2.41 | $7.929 \times 10^{-07}$ | **** |
| DBH (cm) | In_Slow | 54 | 4.63 | 7.61 | 4.38 | 8.94 | 4.77 | - | - |
|  | In_Speed | 54 | 5.35 | 8.00 | 5.12 | 8.75 | 7.35 | 1.0000 | ns |
|  | Out_Slow | 54 | 4.32 | 6.44 | 4.80 | 7.19 | 5.33 | 1.0000 | ns |
|  | Out_Speed | 54 | 4.35 | 6.42 | 3.66 | 7.43 | 6.76 | 1.0000 | ns |
| Crown Width (m) | In_Slow | 38 | 4.88 | 4.91 | 4.78 | 3.51 | 5.09 | - | - |
|  | In_Speed | 38 | 6.54 | 6.54 | 6.12 | 3.88 | 5.34 | $2.818 \times 10^{-02}$ | * |
|  | Out_Slow | 38 | 4.16 | 4.47 | 3.55 | 3.46 | 4.85 | $2.485 \times 10^{-03}$ | ** |
|  | Out_Speed | 38 | 5.84 | 5.89 | 5.67 | 3.64 | 4.46 | 1.0000 | ns |

For DBH estimation, all Wilcoxon comparisons—including In_Slow vs. Out_Slow—returned non-significant p-values (p > 0.05), indicating no meaningful differences in estimation accuracy across scenarios. However, In_Slow still demonstrated reliable performance with a mean error of 4.63 cm, a median of 4.38 cm, and an IQR of 4.77 cm. Out_Slow and Out_Speed showed similar or slightly lower central tendencies (medians of 4.80 cm and 3.66 cm, respectively), but both presented increased spread (IQR = 5.33 cm and 6.76 cm), suggesting more variability in their estimates.

In crown width estimation, the In_Slow scenario showed balanced performance with a median error of 4.78 m, the lowest standard deviation (3.51 m), and a relatively narrow IQR (5.09 m). Although Out_Slow had a slightly lower median error (3.55 m), it exhibited more variability and outliers, compromising its consistency (Figure 13). Wilcoxon test results (see Table 4) indicated that In_Slow significantly outperformed both In_Speed (p = 0.028) and Out_Slow (p = 0.0025). Additionally, Out_Slow and Out_Speed were also significantly different (p = 0.0311), suggesting that increased speed introduces substantial variability in crown width estimation.

Even though In_Slow and Out_Slow perform similarly under ideal conditions (e.g., <10 m distance), In_Slow consistently outperforms Out_Slow in robustness to distance, error consistency, and environmental control. As distance increases, Out_Slow degrades more sharply, especially in geolocation accuracy. In contrast, In_Slow maintains stable error margins and fewer outliers across all structural and positional metrics (Figure 13). Therefore, when considering both attribute accuracy and geolocation reliability, particularly in variable distance conditions, In_Slow emerges as the most dependable and consistent scenario for vegetation measurement using camera-based or remote sensing workflows.

*1.3.3.1 Measurement Error by Object Type*

An object-wise analysis within the optimal In_Slow acquisition scenario revealed distinct error patterns across different object types i.e. tree, pole and other (Figure 14). For height estimation, tree object type exhibited the greatest variability, with errors ranging from –7.92 m to +7.29 m and a mean absolute error (MAE) of ±2.09 m. Poles and other object types had smaller typical errors (MAE = ±0.88 m and ±1.17 m, respectively), indicating a more consistent estimation.

*Table 6: Summary statistics of signed and absolute measurement errors across object types under the In_Slow acquisition scenario. Metrics include the number of observations (n), mean signed error (Mean), mean absolute error (MAE), median, standard deviation (STD), minimum (Min), maximum (Max), and interquartile range (IQR).*

| Error | | n | Mean | MAE | Median | STD | Min | Max | IQR |
|---|---|---|---|---|---|---|---|---|---|
| | Tree | 38 | 0.02 | 2.09 | -0.62 | 2.89 | -7.92 | 7.29 | 2.52 |
| Height (m) | Pole | 16 | -0.38 | 0.88 | -0.62 | 1.1 | -1.63 | 3.2 | 0.64 |
| | Other | 8 | -1.17 | 1.17 | -1.1 | 0.55 | -1.89 | -0.37 | 0.86 |
| DBH (cm) | Tree | 38 | 9.36 | 9.55 | 7.2 | 6.35 | -2.88 | 23.44 | 7.46 |
| | Pole | 16 | 2.85 | 2.98 | 2.94 | 1.65 | -1.04 | 6.1 | 1.31 |
| Crown Width (m) | Tree | 38 | 4.88 | 4.91 | 4.78 | 3.51 | -0.68 | 14.97 | 5.09 |

For DBH, trees also showed greater variability (MAE = ±9.55 cm) compared to poles (MAE = ±2.98 cm). This divergence likely arises from trunk curvature, partial occlusion in oblique views, and irregular bark morphology, which collectively hinder accurate stem diameter extraction from 2D projections.

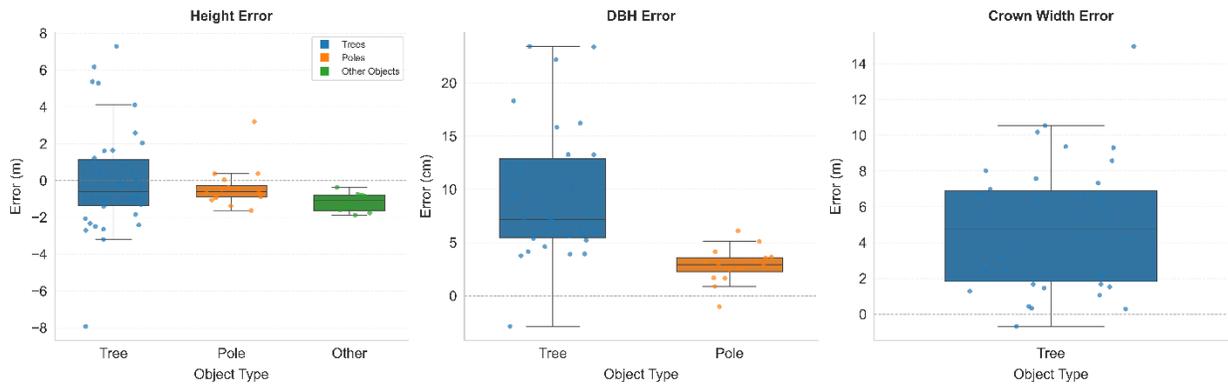

*Figure 14: Object-wise distribution of prediction error for (a) height (m), (b) DBH (cm), and (c) crown width (m) under the optimal In_Slow acquisition scenario. Errors are grouped by object type: Trees, Poles, and Other objects.*

Crown width errors were only present for trees, as pole and other object types lacked the crown, and were found to be MAE of ±4.91 m. The error distribution was positively skewed and exhibited a widespread, suggesting that canopy extent estimation from image data is sensitive to structural complexity and image capture geometry. The magnitude of variability in tree crown estimates underscores the difficulty of deriving reliable lateral canopy measures from single-view image acquisition.

Together, these results demonstrate that while the In_Slow scenario minimizes overall measurement error, vegetation classes such as trees remain more susceptible to structural estimation inaccuracies, particularly in parameters influenced by foliage, occlusion, and form complexity. Engineered objects such as poles showed consistently lower variability across all parameters, highlighting the influence of object morphology on measurement robustness.

Object-level trends in measurement accuracy were further explored by analyzing the relationship between camera-to-object distance, geolocation error, and height error (Figure 15). Objects were indexed sequentially along the capture trajectory, simulating a spatial progression.

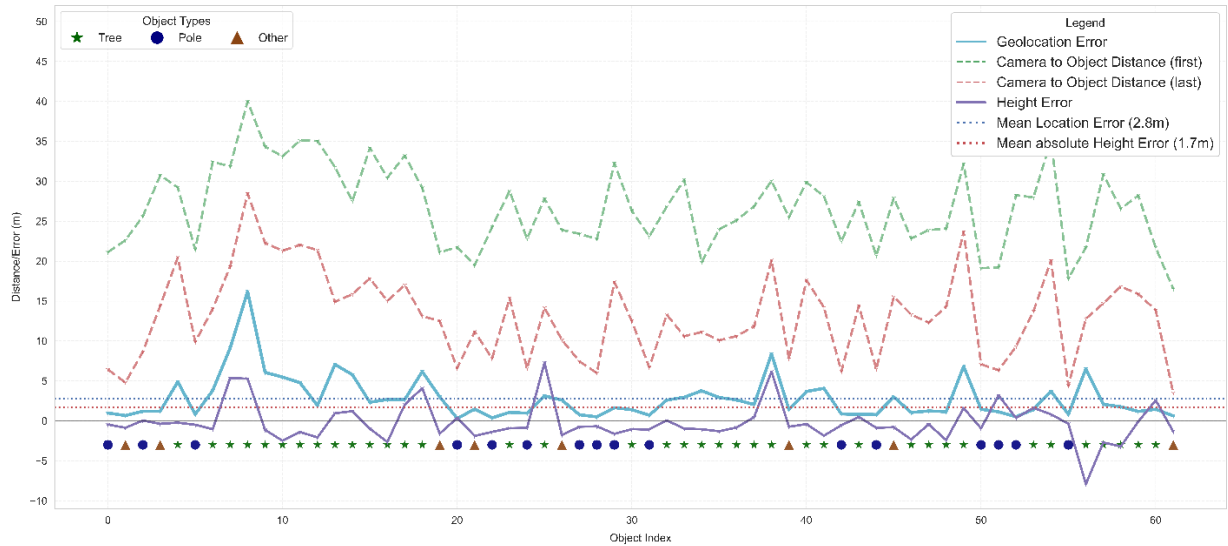

*Figure 15: Object-wise profile of geolocation and structural measurement errors plotted by object index under the In_Slow acquisition scenario. The solid light blue line represents geolocation error, while the dashed lines show the first (red) and last (green) camera-to-object distances. The solid dark purple line indicates a height error. Dashed horizontal lines represent the mean geolocation error threshold (2.8 m) and mean height error threshold (0.2 m). Symbols denote object types: trees (green asterisks), poles (black circles), and other objects (brown triangles).*

Camera distance to the object, plotted for both the first and last frames, varied substantially across the observation set, ranging from ~20 m to over 40 m. As expected, greater distances generally coincided with increased geolocation error, though the relationship was not strictly linear. Peaks in geolocation error (e.g., Object #8, # 39, 50) aligned with steep changes in approach distance and corresponded to localized spikes in height error as well (Figure 15).

Despite the overall low height error mean (-0.2 m), there were localized deviations exceeding ±5 m in a few tree objects, suggesting sensitivity to camera distance or occlusion dynamics. Notably, poles and other objects (denoted by circular and triangular markers) clustered around zero error with little variation, reinforcing prior findings that engineered objects yield more consistent structural estimates.

The horizontal dashed lines indicate the mean geolocation error (2.8 m) and mean height error (-0.2 m), providing a baseline for outlier identification. Several trees exceeded these thresholds simultaneously in both spatial and vertical dimensions, underscoring their susceptibility to compounded error sources.

## 1.4 Discussion

The integration of dashcam video analytics with Deep Learning (DL) presents a transformative opportunity for automated roadside vegetation and infrastructure assessment. Our study demonstrates the feasibility of extracting precise structural and geospatial data from dashcam footage, advancing beyond simple object detection to enable actionable insights for urban ecosystem management. Our study evaluated the performance of camera-based vegetation measurements across four experimental conditions, varying by camera position (interior vs. exterior) and movement speed (slow vs. speed). The findings highlight the sensitivity of structural attribute estimation and geolocation accuracy to both spatial positioning and movement dynamics during data collection.

In evaluating the accuracy of depth estimates across varying distances, it was important to ensure that our measurement approach did not introduce bias from repeated observations of the same object. Although some objects may have appeared in multiple frames, our depth measurements were derived independently from individual frames without temporal aggregation or tracking. Therefore, the observed depth estimation errors, particularly underestimation beyond 15 meters, are unlikely to be influenced by autocorrelation effects and instead reflect model limitations at greater distances. Our XGBoost-based framework achieved robust meter-level depth estimation accuracy (MAE = 4.16 m, $R^2$ = 0.91) in complex urban roadside environments by leveraging domain-aware feature engineering, including square-root scaling and variance-based weighting. While pre-trained deep networks provided initial depth maps, their inability to generalize to our dashcam-specific imaging conditions, especially for distant objects, resulted in significant inaccuracies. Retraining these models would have required extensive labeled datasets from our operational environment, which was impractical. Reformulating the problem as a regression task allowed XGBoost to efficiently correct systematic errors in the pre-trained outputs while maintaining computational efficiency suitable for real-time applications.

This hybrid approach successfully balanced accuracy and scalability, outperforming unrefined DL predictions, mainly enhancing depth estimation accuracy for distant objects (30–50 m). This observation is consistent with the findings in the review by Ezugwu et al. (2024) (Ezugwu et al., 2024), which highlighted the effectiveness of hybrid machine learning and DL frameworks in addressing the limitations of standalone deep models. Our results further demonstrate that ensemble-based refinement can effectively mitigate domain shift artifacts and align with edge-computing paradigms, where lightweight models improve adaptability to deployment environments without compromising precision.

### 1.4.1 Geolocation Analysis

Across all experimental conditions, our framework achieved mean absolute measurement errors ranging from 2.83 m (In_Slow) to 5.01 m (In_Speed), as shown in Table 3. These results indicate strong performance, particularly under low-speed conditions with interior camera placement. Our results are comparable to those of (Krylov et al., 2018)(Krylov et al., 2018) and (Branson et al., 2018)(Branson et al., 2018), who reported positional accuracies of approximately 2 m for poles and traffic lights, and below 2 m for trees using Google Street View imagery. Compared to (Lumnitz et al., 2021)(Lumnitz et al., 2021), who reported measurement errors for trees between 4.31 and 8.55 m across diverse settings, our approach offers improved accuracy. It is important to note that GPS accuracy and image quality are comparatively better in GSV than in our dashcam videos. Hence, these comparisons underscore the effectiveness of our hybrid depth estimation and triangulation framework in achieving reliable object-level measurements across varied operational environments.

Geolocation accuracy was primarily influenced by vehicle speed, with errors rising significantly at higher speeds ($p < 1.2 \times 10^{-5}$; Figure 9). This trend is consistent with findings from (Shami et al., 2024)(Shami et al., 2024) which emphasize the effects of acceleration, deceleration, and GPS signal delay during high-speed vehicle motion. We believe that rapid movement increased frame misalignment and GPS drift which might have caused more error in high-speed scenarios. Interestingly, camera placement (inside vs. outside the vehicle) had a negligible impact on geolocation accuracy, contradicting our initial assumptions. We had anticipated that external mounts would yield better visual clarity than internal mounts as they might be compromised by windshield reflections. However, the XGBoost model likely compensated for possible noise in depth estimation, while the subsequent triangulation process helped mitigate minor geolocation errors. This suggests that model-based corrections and geometry-based methods can effectively balance sensor-induced variability, even under suboptimal capture conditions.

Also, geolocation accuracy is highly dependent on the distance between the object and the last observed camera frame. Objects within 10m were located with high precision across all conditions, highlighting the reliability of close-range detection as shown in Figure 12. However, accuracy declined at greater distances, particularly in the In_Speed conditions, where increased motion and reduced object clarity likely contributed to compounded triangulation errors. These findings are consistent with the work of (Lumnitz et al., 2021)(Lumnitz et al., 2021) and (Shami et al., 2024)(Shami et al., 2024), who reported increased geolocation uncertainty with growing camera-to-object distances. Additionally, we found that the distance from the last camera frame had a stronger correlation with geolocation accuracy than the first, reinforcing the importance of final-frame proximity as a reliable predictor in monocular triangulation. This suggests that both spatial proximity and vehicle dynamics are key factors influencing geolocation performance in monocular image-based systems.

### 1.4.2 Tree Structure Analysis

The comparative evaluation across scenarios revealed that the video taken from inside the vehicle at slower speed (<40 kmph) (In_Slow condition) consistently yielded more accurate and stable vegetation measurements along with low geolocation error. While Out_Slow occasionally matched In_Slow in central tendency, its increased error variability, especially as object distance increased, highlights it may be sensitive to vibration. The lack of significant differences in DBH accuracy across scenarios further supports the notion that trunk-based features are less affected by capture dynamics. However, for more complex attributes like crown width and height, In_Slow offered a clear advantage in terms of robustness. Also, one potential confounding factor in our scenario comparison is the difference in camera height between the In_Slow (1.2 m) and Out_Slow (0.9 m) conditions. While seemingly minor, this variation can affect the projection geometry of objects in monocular video, particularly for vertical attributes such as height and crown width (Yan & Huang, 2022). A higher viewpoint (inside mounting) may offer a slightly better angle for depth estimation and reduce foreground occlusion, especially in sloped or cluttered environments. This could partially explain the improved consistency observed in the In_Slow condition, beyond the effects of speed or camera placement alone. Although this height difference was not explicitly controlled in our design, it highlights the importance of standardized mounting height in future deployments aimed at accurate structural measurements.

In our best scenario, i.e. In_Slow, structural analysis revealed distinct error patterns across object types. Trees exhibited higher variability in height, DBH, and crown width (Table 6), likely due to occlusion, foliage complexity, and irregular morphology. These challenges mirror those reported in forestry remote sensing, where 2D imagery struggles to capture 3D canopy structures (Deng et al., 2023)(Deng et al., 2023). In contrast, utility poles which are geometrically simpler and vertically mostly uniform were predicted with higher consistency, showing much lower mean absolute height and DBH error. This suggests that our methodology can capture the linear measurement, however, could not account for undulation due to the limitation of the 2D perspective.

### 1.4.3 Limitations and Future Directions

While our study demonstrates the feasibility of using monocular dashcam imagery for roadside vegetation and infrastructure assessment, several limitations must be acknowledged. One of the key constraints lies in the spatial resolution and accuracy of geolocation data. The built-in GPS in our dashcam has meter-level uncertainty and provides geotags for only one frame per second. This limited the number of usable frames for each object, thereby reducing the temporal density of observations and constraining the triangulation process. Increasing the spatial metadata, either through higher-frequency GPS logging or integration with external GPS/IMU systems, would increase frame correspondence and improve geolocation precision.

Another important limitation was the inherent distortion in the captured images. As the dashcam was not calibrated for distortion removal, minor warping or perspective distortion may have affected depth estimation and structural measurements, particularly toward the edges of the field of view. While we did not aim to remove all distortions, since we wanted to develop a method robust enough for noisy, citizen-science-style data, these distortions nonetheless introduced a degree of spatial uncertainty. Future work could explore adaptive distortion correction methods that balance robustness with generalizability.

Although our dashcam recorded at higher resolution, distant objects often appeared too small or blurry for reliable annotation, limiting the dataset to objects within a moderate range. This prevented the evaluation of object detection and measurement performance for far-field vegetation, which may behave differently in terms of depth estimation and visibility. Utilizing zoom-capable or multi-sensor imaging systems may help extend detection capacity to longer distances. Additionally, while our methodology performed well for structured objects like utility poles, performance degraded for complex vegetation due to occlusion, irregular morphology, and canopy overlap. These challenges mirror those in broader urban forestry, where 2D imagery struggles to capture true 3D structure(Gougeon & Leckie, 2006; Takahashi et al., 2012). Expanding the framework to support species-level classification and volumetric canopy modeling would significantly enhance ecological relevance.

Finally, the current workflow remains semi-automated, with manual annotation and selection of keyframes. Automating object detection, multi-frame tracking, and triangulation will be essential for scaling the methodology to continuous monitoring applications. Future development should focus on integrating real-time processing pipelines capable of detecting, localizing, and measuring features with minimal human intervention. These limitations outline key areas for improvement but also reinforce the potential of dashcam-based video analytics as a scalable, low-cost solution for roadside monitoring, particularly for citizen-collected data context.

## 1.5 Conclusion

This study bridges a critical gap in roadside infrastructure and vegetation assessment by demonstrating the viability of dashcam video analytics for precise, scalable urban ecosystem monitoring. It demonstrates the feasibility and effectiveness of using dashcam-based monocular video, combined with a hybrid DL and machine learning framework, for accurate roadside vegetation and infrastructure assessment. By integrating pre-trained monocular depth estimation with XGBoost-based refinement and then triangulation, we achieved reliable geolocation and structural measurement accuracy, with MAE as low as 2.83 m under optimal conditions. Our results highlight the influence of vehicle speed, object distance, and viewpoint height on measurement performance. The In_Slow scenario, characterized by low speed and interior mounting, consistently yielded the most accurate and stable results across both geolocation and structural attributes.

Notably, the framework showed strong performance in localizing and measuring geometrically simple objects like utility poles, while greater variability was observed for complex vegetation such as trees—reflecting the inherent limitations of 2D imaging for irregular 3D structures. The minimal impact of camera mounting location and the compensatory role of model-based correction suggest that thoughtful integration of geometry-aware methods can mitigate common sensor-related limitations. Additionally, the findings emphasize the need for standardized data capture protocols, such as consistent camera height and controlled speed, to enhance measurement robustness.

While the approach demonstrated strong performance, certain limitations such as GPS resolution and image distortion highlight opportunities for refinement. Future work should focus on enhancing automation, extending species-level analysis, and improving system robustness under varied field conditions. Overall, this work underscores the potential of lightweight, scalable video analytics for automated roadside monitoring, and offers a practical foundation for future applications in urban forestry, utility management, and intelligent transportation systems.

**CRediT authorship contribution statement**

**Durga Joshi**: Conceptualization, Methodology, Data Curation, Software, Validation, Formal analysis, Writing - original draft; **Chandi Witharana**: Conceptualization, Methodology, Supervision, Writing - original draft, Writing - review & editing; **Robert Fahey**: Supervision, Writing - review & editing; **Thomas Worthley:** Supervision, Writing - review & editing; **Zhe Zhu:** Supervision, Writing - review & editing; **Diego Cerrai**: Supervision, Writing - review & editing.

**Declaration of Competing Interest**

The authors declare that they have no known competing financial interests or personal relationships that could have appeared to influence the work reported in this paper.

**Acknowledgement**

We would like to thank the Eversource Energy Center at the University of Connecticut (UConn), USA, for their support throughout this work. We acknowledge the Department of Natural Resources and the Environment, UConn for facilitating field data collection by providing departmental vehicles.

Kowe, P., Mutanga, O., & Dube, T. (2021). Advancements in the remote sensing of landscape pattern of urban green spaces and vegetation fragmentation. *International Journal of Remote Sensing, 42*(10), 3797–3832.

Krylov, V. A., Kenny, E., & Dahyot, R. (2018). Automatic discovery and geotagging of objects from street view imagery. *Remote Sensing, 10*(5), 661.

Li, X. (2021). Examining the spatial distribution and temporal change of the green view index in New York City using Google Street View images and deep learning. *Environment and Planning B: Urban Analytics and City Science, 48*(7), 2039–2054.

Li, X., Ratti, C., & Seiferling, I. (2018). Quantifying the shade provision of street trees in urban landscape: A case study in Boston, USA, using Google Street View. *Landscape and Urban Planning, 169*, 81–91. 10.1016/j.landurbplan.2017.08.011

Li, X., Zhang, C., Li, W., Ricard, R., Meng, Q., & Zhang, W. (2015). Assessing street-level urban greenery using Google Street View and a modified green view index. *Urban Forestry & Urban Greening, 14*(3), 675–685.

Liang, J., Gong, J., Zhang, J., Li, Y., Wu, D., & Zhang, G. (2020). GSV2SVF-an interactive GIS tool for sky, tree and building view factor estimation from street view photographs. *Building and Environment, 168*, 106475.

Lumnitz, S., Devisscher, T., Mayaud, J. R., Radic, V., Coops, N. C., & Griess, V. C. (2021). Mapping trees along urban street networks with deep learning and street-level imagery. *ISPRS Journal of Photogrammetry and Remote Sensing, 175*, 144–157. 10.1016/j.isprsjprs.2021.01.016

Lyft. (2022). *Recording device policy, Lyft.* https://help.lyft.com/hc/en-us/articles/115012923127-Safety-policies (2022)). Retrieved 4/13/2022, from

Marinov, D., & Karapetyan, D. (2019). Hyperparameter optimisation with early termination of poor performers. Paper presented at the *2019 11th Computer Science and Electronic Engineering (CEEC),* 160–163.

Nawar, N., Sorker, R., Chowdhury, F. J., & Rahman, M. M. (2022). Present status and historical changes of urban green space in Dhaka city, Bangladesh: A remote sensing driven approach. *Environmental Challenges, 6*, 100425.

Ow, L. F., & Ghosh, S. (2017). Urban cities and road traffic noise: Reduction through vegetation. *Applied Acoustics, 120*, 15–20. https://doi.org/10.1016/j.apacoust.2017.01.007

Randrup, T. B., McPherson, E. G., & Costello, L. R. (2001). A review of tree root conflicts with sidewalks, curbs, and roads. *Urban Ecosystems, 5*, 209–225.33